\documentclass[a4paper,conference]{IEEEtran}
\usepackage{graphicx}
\usepackage{makecell} 
\usepackage{multirow}
\usepackage{amsmath}
\usepackage{arydshln}

\ifCLASSINFOpdf
\else
\fi
\usepackage{amsmath,amssymb}
\newcommand{\argmin}[1]{\underset{#1}{\operatorname{arg}\,\operatorname{min}}\;}
\newcommand{\etal}{\textit{et al. }}

\begin{document}
%
\title{Online Deep Clustering with Video Track Consistency}

\author{\IEEEauthorblockN{Alessandra Alfani, Federico Becattini, Lorenzo Seidenari, Alberto Del Bimbo}
\IEEEauthorblockA{University of Florence}
}


%


\maketitle

\begin{abstract}
Several unsupervised and self-supervised approaches have been developed
in recent years to learn visual features from large-scale unlabeled datasets.
Their main drawback however is that these methods are hardly able to recognize visual features of the same object if it is simply rotated or the perspective of the camera changes.
To overcome this limitation and at the same time exploit a useful source of supervision, we take into account video object tracks. Following the intuition that two patches in a track should have similar visual representations in a learned feature space, we adopt an unsupervised clustering-based approach and constrain such representations to be labeled as the same category since they likely belong to the same object or object part.
Experimental results on two downstream tasks on different datasets demonstrate the effectiveness of our Online Deep Clustering with Video Track Consistency (ODCT) approach compared to prior work, which did not leverage temporal information.
In addition we show that exploiting an unsupervised class-agnostic, yet noisy, track generator yields to better accuracy compared to relying on costly and precise track annotations.
\end{abstract}


%
\IEEEpeerreviewmaketitle

\section{Introduction}
Convolutional Neural Networks (CNNs) are being widely used as the main framework in many computer vision applications
due to their great ability to learn different levels of general visual features.
Over the past 10 years, their performance has continued to improve, thanks to complex architectures and large-scale datasets. 
The success of CNNs is, in fact, highly dependent on their capabilities to model highly non-linear patterns and the amount of training data available. The deeper and more expressive the network is, the larger the required dataset will be to effectively train such model.

Models trained on large-scale, fully-supervised and diverse image datasets, such as ImageNet~\cite{imagenet} and YFCC100M~\cite{2016}, have been demonstrated to capture common and generic visual features, making them also ideal as a starting point for diverse subsequent learning tasks.
According to this view, collecting and annotating in a precise way larger and more diversified datasets to be used in supervised machine learning seems a natural approach to go forward. However, on the other hand, this approach is extremely costly, time consuming and requires a massive amount of manual annotations.
As a consequence, unsupervised learning has recently received attention in the field of machine learning and methods for clustering, dimensionality reduction, and density estimation are commonly used in computer vision applications~\cite{Joulin2010DiscriminativeCF, 868688, becattini2017indexing, Csurka04visualcategorization}. 

The goal of unsupervised representation learning is indeed to learn transferable image or video representations without the need for manual annotations~\cite{doersch2016unsupervised, pathak2016context, zhang2016colorful, noroozi2017unsupervised, berlincioni2021multiple, donahue2017adversarial, de2022disentangling}. 
Clustering-based representation learning approaches, which jointly optimize clustering and feature learning, stand out as a potential trend in this field \cite{Huang_2016_CVPR, xie2016unsupervised, yang2016joint, caron2019deep, Caron_2019_ICCV, zhan2020online}.

In this work we focus the attention on two unsupervised clustering-based learning methods, DeepCluster (DC) \cite{caron2019deep} proposed by Caron \textit{et al.} and Online Deep Clustering (ODC) \cite{zhan2020online}  proposed by Zhan \textit{et al.}. Both these methods alternate between deep feature clustering and CNN parameters update, with the difference that ODC proposed to decompose the clustering procedure into mini-batch-wise label updates and incorporate these updates into network update iterations.

Despite the success of such clustering-based methods, they are mostly focused on still images and do not exploit additional sources of supervision such as temporal consistency in videos.
Following this idea, we propose a novel unsupervised approach to exploit temporal information from videos, allowing the network to use tracks proposals as a form of self-supervision. Our method, dubbed Online Deep Clustering with Video Track Consistency (ODCT), is fully unsupervised and can be trained starting from any video database, without any prior knowledge regarding the nature of their content. To this end we leverage an unsupervised object discovery approach that extracts class-agnostic track proposals and exploit such data to constrain a feature clustering phase and generate temporal coherent pseudo-labels.

The main contributions of our work are the following:
\begin{itemize}
    \item We present a fully unsupervised, class-agnostic method exploiting video temporal consistency of object tracks using feature clustering to generate pseudo-labels.
    \item We show how adding constrains on the clustering phase we can account for intra-track variability and obtain a more effective supervision signal.
    \item We test our model on two downstream tasks on different datasets, obtaining significant improvements compared to prior work which leverages only still images.
\end{itemize}

\section{Related Work}
Unsupervised and Self-Supervised learning are used in representation learning methods to generate feature representations merely from images, without the need for time-consuming semantic annotations. 
Self-Supervised methods, in particular, employ pre-designed pretext tasks and need no labeled data in order to train the weights of a convolutional neural network. Instead, the visual features are learned by minimizing the pretext task's objective function.

Several pretext tasks have been proposed in the literature and each of them was carefully designed to use various cues from images or videos. 
Noroozi \etal  proposed to learn image representations by solving Jigsaw puzzles~\cite{noroozi2017unsupervised}. \textit{RotNet}~\cite{gidaris2018unsupervised} relies on predicting image rotations. Other approaches exploit temporal or image context prediction. Doersch~\etal leverage image context~\cite{doersch2016unsupervised} while Misra~\etal seek supervision via temporal order verification~\cite{misra2016shuffle}.

A different line of work relies instead on generative or completion tasks. Zhang \etal learns representations via image colorization~\cite{zhang2016colorful}; Pathak~\etal 
use image inpainting instead~\cite{pathak}; Tulyakov relies on video generation with GANs \cite{tulyakov2017mocogan} to learn features.

Several strategies for combining multiple cues have been proposed recently, Doersch \textit{et al.} in \cite{doersch2017multitask} proved that combining diverse pretext tasks to simultaneously train a single trunk network, using a multi-task objective function, can improve the overall performance.
However, these methods are domain-specific and require expert knowledge to properly design a pretext task that may result in transferable features.

Moreover, many approaches, such as inpainting or colorization, have the disadvantage of learning features on modified images, which may impair generalization to unmodified ones. Colorization, for example, takes a grayscale image as input, so the network cannot learn to extract color information, which might be useful for other tasks.



Clustering is often employed as a means to generate supervision via pseudo-labels~\cite{caron2019deep,noroozi2018boosting, zhan2020online}. Noroozi \etal propose to use k-means after a jigsaw based pretraining~\cite{noroozi2018boosting}, a simple classifier is then trained from scratch on the gathered pseudo-labels. Caron \etal embed the clustering step in the training loop, thus updating the pseudo-labels during all training\cite{caron2019deep}. Zhang \etal improved over~\cite{caron2019deep} by updating labels continuously rather than in a pulsating manner, enabling the representations to evolve steadily.

Some works try to avoid the use of top-down clustering. Instead of generating synthetic classes via grouping, Wu \etal propose to learn a representation capturing the similarity among instances by requiring learned features to be discriminative of individual instances~\cite{wu2018unsupervised}. Huang \etal propose a neighborhood discovery approach based on a divide-and-conquer strategy~\cite{huang2019unsupervised}. The proposed method is able to find class consistent neighbourhoods anchored to individual training samples which result in highly compact sample clusters.

To get the best of both worlds, Vangan \etal~\cite{vangansbeke2020scan}, combine a non-clustering based pre-training step based on a pretext task. The obtained features are used as a prior in a learnable clustering approach. This allows to remove the need for clustering to rely on low-level features.

In this work we follow the clustering-based pre-training strategies, in particular building upon the Online Deep Clustering (ODC) approach~\cite{zhan2020online}. Differently from previous work though, we exploit videos to perform the pre-training stage and leverage temporal coherence as a source of supervision. By extracting object tracks from videos we are able to formulate a novel constrained clustering which incrementally builds pseudo-labels attempting to assign samples belonging to the same track to the same cluster in the feature space.

\section{Online Deep Clustering with Video Track Consistency}
\label{sec:method}
	
	\begin{figure}
		\centering
		\includegraphics[width=\linewidth]{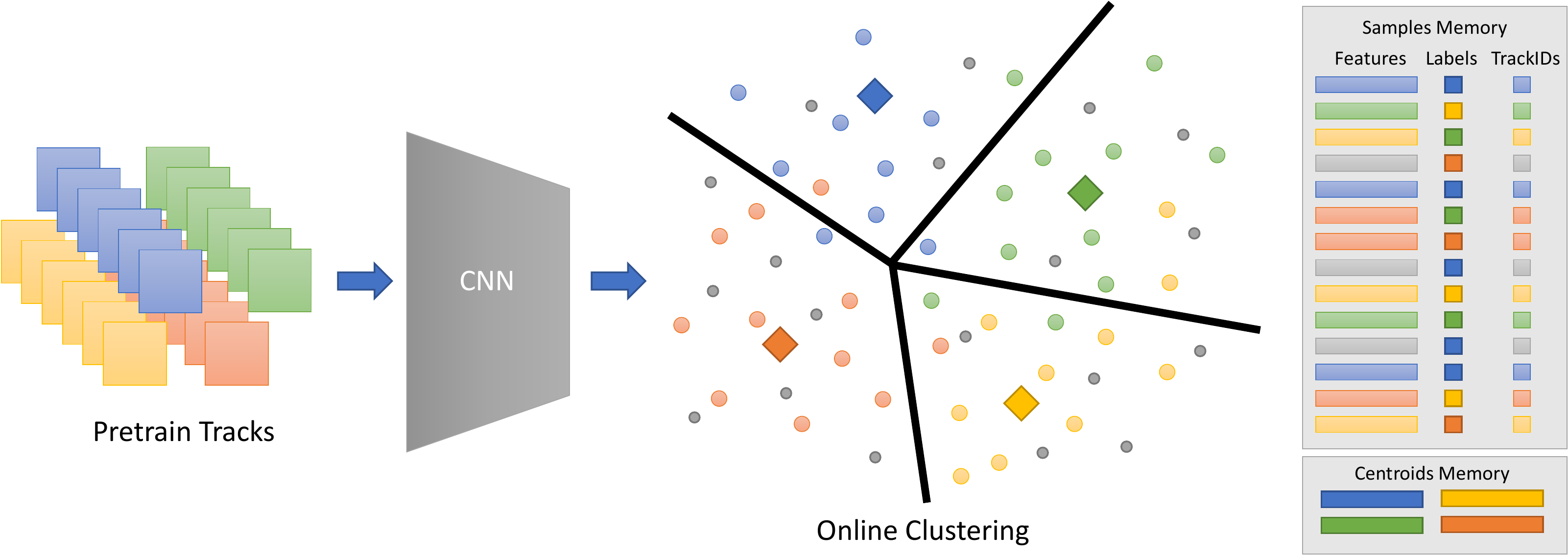}
		\caption{Pseudo-labels are generated during training with online clustering. Track identities are stored in an external memory to constrain the clustering stage and make features of the same track belong to the same cluster.}
		\label{fig:model}
	\end{figure}
	
	In this paper we extend Online Deep Clustering~\cite{zhan2020online} by exploiting video temporal coherency as a self-supervised source for constraining the clustering stage (Fig. \ref{fig:model}).
	
	Given a video $V=f_0,...,f_{N-1}$, where $f_i$ is the $i$-th frame, we define a track as a sequence of $K$ bounding boxes $b_i^k$, $k=0,...,K-1$. A track spatially encloses in its boxes the same object across a set of adjacent frames $[f_j, f_{j+K}]$. We do not assume to have any information regarding the class of such object, but we assume to have access to an oracle, capable of yielding tracks associated with different objects in a video.
	
	Our goal is to train a classifier using pseudo-labels that are consistent at track level. The advantage of using tracks is that they naturally provide a supervision signal that links different views of the same object, therefore accounting for intra-class variability stemming from pose and viewpoint.
	
	ODC originally relies on two external memories, the Samples Memory and the Centroids memory. Such memories are used to store features for training samples and their correspondence with the centroids of each cluster, i.e. the pseudo-labels used for training the model. We extend this formulation by adding a Track Memory where we store a track identifier for each sample.
	As in standard ODC, we populate the Samples Memory by extracting features from a randomly initialized CNN. Differently from \cite{zhan2020online}, however, we consider each bounding box in a frame as a different sample. Therefore, features are extracted from image crops representing objects instead of from whole images.
	For the initial clustering step, we draw a representative at random from each track and perform K-means. We store the centroids in the Centroids Memory, propagating the cluster of each representative to all the other samples in the track.
	After the initialization step, we train the network following five steps:
	
		(i) \textit{Network Forward Propagation}: given a batch of samples $x_i$, the network extracts compact feature vectors $f_{\theta}(x_i)$ and a prediction $\tilde{y}_i$.
		
		(ii) \textit{Network Backward Propagation}: pseudo-labels $p_i$, associated with each sample, are retrieved from the Samples Memory and are used to compute the loss. The parameters of the CNN are updated accordingly after gradient computation.
		
		(iii) \textit{Weight Distances}: for each feature $f_{\theta}(x_i)$ extracted in the current batch, we retrieve from the Samples Memory all the features belonging to samples $x_j$ in the same track $F_m(x_j)$. Let $x^{\prime}$ be the sample in the track with the closest feature vector $F_m(x^{\prime})$ to $f_{\theta}(x_i)$. We define weight coefficients $d_j$, that will be used to update the Samples Memory, as
		$d_j = \frac{d(x_i, x^{\prime})}{d(x_i, x_j)}, \forall x_j \in T$
		
		(iv) \textit{Samples Memory Update}: each sample in the current batch is used to update the Samples Memory exploiting the weight coefficients $d_j$:
		\begin{equation}
		F_m \leftarrow m\frac{f_\theta(x)}{\parallel f_\theta(x) \parallel _2} + (1-m)\frac{\sum_{j}d_jF_m(x_j)}{\sum_{j}d_j}
		\label{odcteq}
		\end{equation}
		where $m\in(0, 1]$ is a momentum coefficient.
		At the same time, each sample's pseudo-label is updated by finding its nearest centroid:
		\begin{equation}
		\argmin{p\in\{1..C\}}{\left \| F_m-C_{p} \right \|^2_2},
		\label{kmeansew}
		\end{equation}
		where $C_p$ indicates the centroid feature of class $p$ in the Centroids Memory.
		
		(v) \textit{Centroids Memory Update}: every $k$-th iteration the centroids are updated as the mean of the feature vectors assigned to each cluster.
	
	As in \cite{zhan2020online}, to avoid trivial solutions where all samples are assigned to a single clusters, we prevent small clusters to become empty.
	Let $C_s$ be the set of clusters with size less than a certain threshold. For $c \in C_s$, we start by assigning samples in $c$ to the nearest centroids belonging to the remaining clusters $C_n = C \setminus C_s$, thus making $c$ empty.
	Then, the largest cluster $c_{max} \in C_n$ is partitioned into two sub-clusters using \textit{K-Means}.
	This process is repeated until the set of small clusters $C_s$ is empty.
	
	\section{Track Generation}
	\label{sec:track}
    To train our model, we assumed to have access to a set of video tracks. For the purpose of our work, these can be provided by some source of track oracle yielding linked bounding boxes in video frames. In the following Sec. \ref{sec:exp} we will discuss experiments performed using ground truth tracks.
	Nonetheless, since we operate under an unsupervised data-regime, we also rely on an object discovery pipeline, derived from \cite{cuffaro2016segmentation}, capable of producing a set of class-agnostic tracks which can be exploited to train our model.
	The idea of such approach is to track spatio-temporal consistency in frame-wise object proposals. We use EdgeBoxes \cite{Zitnick2014EdgeBL} to generate proposal boxes for each frame. These boxes are likely to enclose an object based on low level image characteristics, such as edges, which provide an \textit{objectness} score for regions of interest. Note that this method does not require any class-specific training, thus can be applied to discover any kind of object or part of it.
	As in \cite{cuffaro2016segmentation}, we then use a greedy box-association tracker to link proposals across frames based on their spatial consistency through time. To reduce mismatches, we register each frame onto the next by shifting boxes using optical flow.
	More formally, we can define a track as a succession of bounding boxes $b^k_i$, for which the intersection over union (IoU) between two boxes of consecutive frames, $b^n_i$, $b^m_{i+1}$, is greater than a threshold $\theta_\tau$.
	We can define as well the track score $S_{t_j}$ for track $t_j$, as the mean objectness score of its boxes
	$S_{t_j} = \frac{\sum_{i \in t_j}s_i^k}{|t_j|}$
	where $|t_j|$ indicates the total number of boxes in track $t_j$.
	This naturally leads to a track ordering.
	
	An issue with matching bounding boxes is that there might be missing frames and therefore track fragmentation due to occlusion or appearance changes.
	As in~\cite{Zitnick2014EdgeBL} we use a Time to Live counter (TTL), which represents the maximum number of frames without match before truncating a track.
	Finally, we apply a post-processing step to remove the tracks that are unlikely to represent objects. Particularly, we consider only tracks with a length between $l$ and $L$, as we believe that shorter, more consistent tracks are preferable to longer tracks that are likely to be subject to errors, due to static background, noise, object occlusion or camera movement. $l$ and $L$ are empirically set to 40 and 100, respectively.
	In order to to retain only tracks that are likely to represent objects, we retain only the first 15 tracks in each snippet.
	
\section{Experiments}
\label{sec:exp}

In this Section, we present experimental results.
In addition to our \textit{ODC with Video Track Consistency} (ODCT) model, for the purpose of evaluation we report results from two additional methods.
The first is vanilla ODC \cite{zhan2020online}, trained on the same video data as ODCT. This does not integrate temporal knowledge from video snippets in any way and relies purely on clustering visually similar items.
Then we propose a simple ODC variant with our modified \textit{K-Means} procedure used for initialization, which constrains crops belonging to the same track to be assigned to the same cluster. We will refer to this model as ODC$_{TrackInit}$.


We apply a two-step training procedure, commonly adopted for unsupervised learning \cite{caron2019deep, zhan2020online}:
(i) First, we perform unsupervised pre-training using the unsupervised pretext methods on a pre-training dataset. Usually, such dataset is large yet with limited or no annotation.
(ii) We then perform transfer learning by fine-tuning the model on some downstream task, i.e. harder tasks with access to some form of supervision.

\begin{figure*}[t]
\centering
\begin{minipage}{0.55\textwidth}
\centering
    \includegraphics[width=\textwidth]{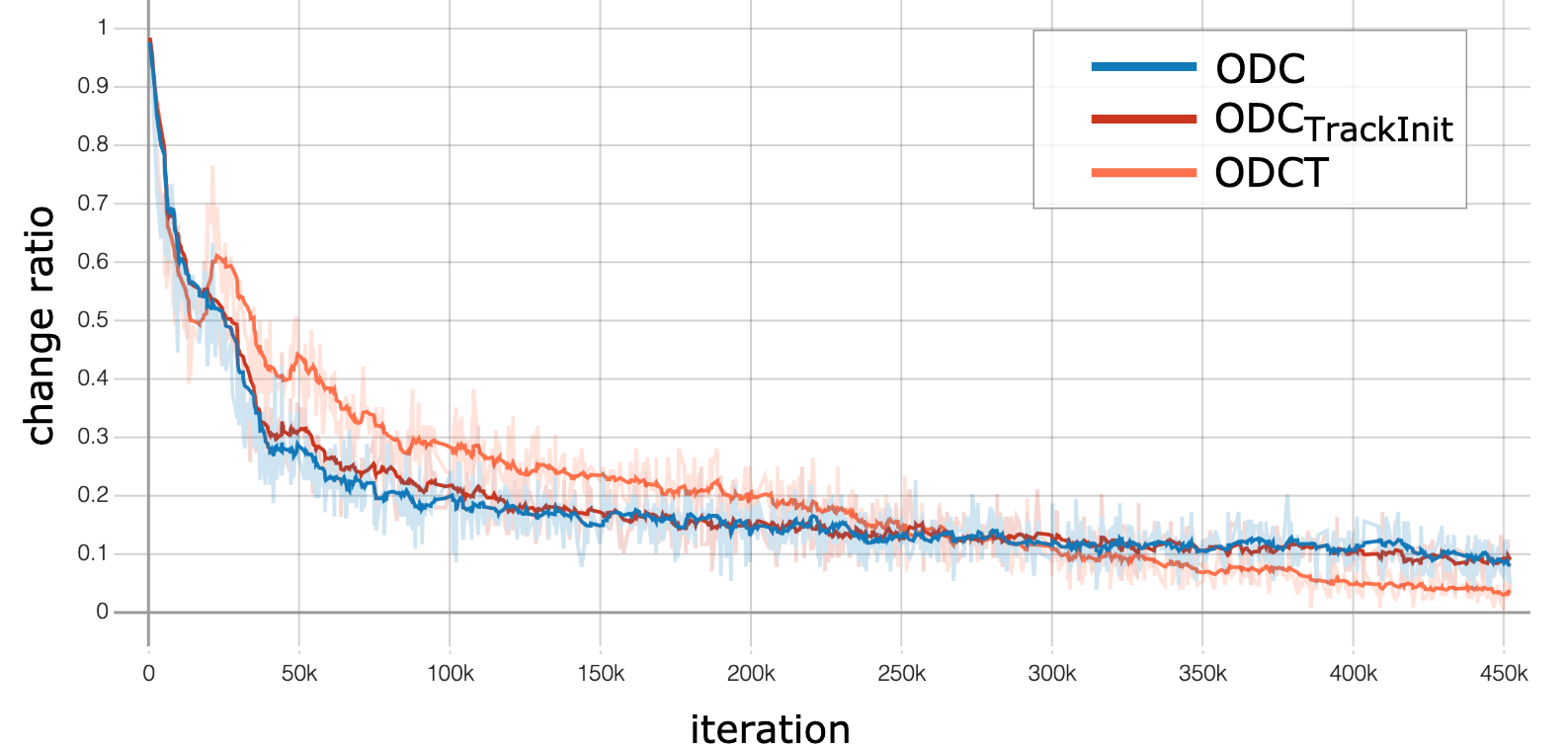}
    \caption{Change ratio for sample-cluster assignments during training.}
    \label{fig:changeratio}
\end{minipage}
\begin{minipage}{0.35\textwidth}
\centering
    \includegraphics[trim={10px 0 40px 0},clip,width=\textwidth]{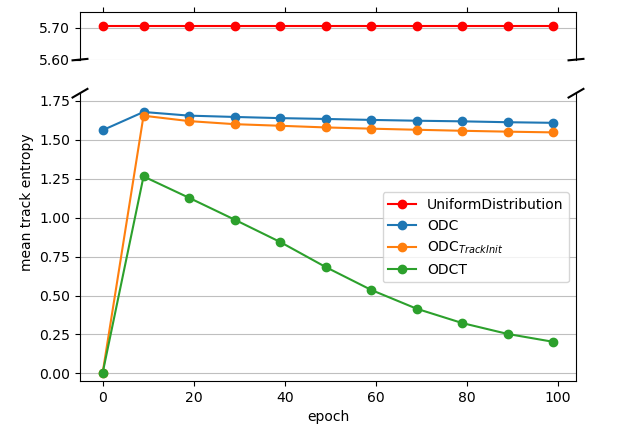}
    \caption{Mean track entropy for cluster assignments during training. We also report the entropy for a Uniform Distribution over clusters as an upper bound.}
    \label{fig:meantrackentropy}
\end{minipage}
\end{figure*}

\subsection{Datasets and Data Preprocessing}

We conduct our pre-training experiments on the ImageNet VID 2015 dataset \cite{ILSVRC15}, a dataset designed for object tracking and object detection.
It includes videos of 30 basic object categories, subset of the 1,000 ILSVRC-2012\cite{imagenet} classes.
In total, ImageNet VID consists of 3,862 snippets for training, 555 snippets for validation, and 937 snippets for testing. All snippets include 56 to 458 frames of images, with a median frame rate and duration of 29\textit{fps} and 12\textit{s}, respectively.
Each frame is annotated with labels indicating the presence of the 30 object classes and their corresponding bounding boxes.
The dataset comprises a total of 7,857 ground truth tracks for training. In our experiments we will initially discard such annotations in order to work in a fully unsupervised regime, but we will also train our model using ground truth tracks as a control experiment in Sec. \ref{sec:gt}.
As downstream tasks we follow commonly adopted benchmarks as in \cite{zhang2016colorful, caron2019deep, zhan2020online} and evaluate our model on ImageNet \cite{imagenet} and Pascal VOC 2007 \cite{pascal-voc}.

We scale all frames to the same maximum dimension of 600px.
As described in Sec. \ref{sec:track}, we perform the object proposal phase with EdgeBoxes. To reduce noise, we limit the number of proposals per frame to 300, before applying non-maximal suppression (NMS) to select the best bounding box for each object.
In the box tracking phase, we set the Time To Live counter (TTL) to 3 frames and the IoU threshold for bounding box matching to 0.35. Furthermore, we removed bounding boxes with an area smaller than 10,000 square pixels.
We collected 57,879 tracks, each of which has been subsampled by picking 10 frames, equally spaced in time, for a total of 578,790 training samples.
Object crops are resized to a resolution of 224x224 and data augmentation is applied including random flipping, rotation (±2$^\circ$) and color jittering.
In addition, we randomly convert images to grayscale with a probability of 0.2. Applying random color jitter and grayscale transformations on the training samples, discourages the network from exploiting trivial information from color to group samples snippet-wise.

\subsection{Training Details}
\label{sec:trainingdetails}

We employ ResNet-50 as model backbone and introduce a non-linear head to reduce hidden size dimensionality for storing features in the external memories. The head is composed of a 512-dimensional fully connected layer, followed by batch normalization, another 256-dimensional fully connected layer with relu activation and dropout. The head layer is removed for downstream tasks.

All the models are trained from scratch on a single GeForce RTX 2080 Ti GPU with mixed precision training for efficiency.
The batch size and the learning rate are respectively set to 128 and 0.015 for 100 epochs using the SGD optimizer with momentum set to 0.5.
The threshold to identify small clusters is set, as in ODC, to 20. Zhan \textit{et al.} \cite{zhan2020online} proved that changing this threshold does not affect the results significantly, as long as it does not exceed the average number of samples in a cluster.
Regarding the Centroids Memory update, we perform it every 10 iterations to balance learning efficacy and efficiency.
Following Caron \textit{et al.} \cite{caron2019deep}, we use a number of clusters equal to 10 times the number of annotated categories in the dataset (K=300). However, as an ablation study, we also trained our model using K=30 and K=1000.



\begin{table*}[!htb]
    \caption{Top-1 classification accuracy on ImageNet (left) and mAP on Pascal VOC 2007 (left).  ODC$^*$\cite{zhan2020online}: ODC pre-trained on ImageNet. \textit{ImageNet labels}: fully supervised ResNet-50 pre-training on ImageNet. \textit{Random}: ResNet-50 with random weights. All other models are pre-trained on ImageNet VID without supervision.}
    \label{tab:imagenetpascal}
    \begin{minipage}{.5\linewidth}
      \centering
      \resizebox{\textwidth}{!}{
        \begin{tabular}{llccccc}
    
    \hline
    \textbf{ImageNet} & \textbf{Method} & Stage1 & Stage2 & Stage3 & Stage4 & Stage5   \\ \hline
     & ImageNet labels\cite{zhan2020online} & 15.18 & 33.96 & 47.86 & 67.56 & 76.17\\
     & Random\cite{zhan2020online} & 11.37 & 16.21 & 13.47 & 9.07 & 6.54\\
     & ODC$^*$\cite{zhan2020online} & 14.76 & 31.82 & 42.44 & 55.76 & 57.70\\ \hline
    \multirow{3}{*}{\textit{K = 30}}
        & ODC & 10.63 & 21.74 & 23.01 & 22.86 & 16.65\\
        & ODC$_{TrackInit}$ & 10.61 & 22.27 & 24.10 & 24.07 & 17.34\\
        & ODCT & \textbf{10.91} & \textbf{23.50} & \textbf{27.37} & \textbf{30.26} & \textbf{23.38}\\
      \hline
    \multirow{3}{*}{\textit{K = 300}}
        & ODC & 9.95 & 21.92 & 23.70 & 23.35 & 16.74\\
        & ODC$_{TrackInit}$ & 10.39 & 21.74 & 25.56 & 25.86 & 19.20\\
        & ODCT & \textbf{11.02} & \textbf{24.54} & \textbf{28.06} & \textbf{32.23} & \textbf{26.00}\\
       \hline
    \multirow{3}{*}{\textit{K = 1000}}
    & ODC & 11.14 & 21.25 & 23.65 & 23.14 & 16.80\\
    & ODC$_{TrackInit}$ & 11.39 & 23.11 & 24.82 & 25.93 & 19.53\\
    & ODCT & \textbf{11.44} & \textbf{24.79} & \textbf{28.72} & \textbf{32.95} & \textbf{26.04}\\
        
    \hline
    \end{tabular}
    }
    \end{minipage}%
    \begin{minipage}{.5\linewidth}
      \centering
        \resizebox{\textwidth}{!}{
        \begin{tabular}{llccccc}
    \hline
    \textbf{VOC07} & \textbf{Method} & Stage1 & Stage2 & Stage3 & Stage4 & Stage5 \\ 
\hline
     &ImageNet labels\cite{zhan2020online} & 26.84 & 47.56 & 58.94 & 78.94 & 87.17\\
     & Random\cite{goyal2019scaling} & 9.60 & 8.30 & 8.10 & 8.00 & 7.70\\
     & ODC$^*$\cite{zhan2020online} & 27.33 & 46.16 & 56.22 & 68.06 & 78.42\\
    \hline
    \multirow{3}{*}{\textit{K = 30}}
        & ODC & 24.80 & 39.36 & 40.02 & 36.72 & 30.85\\
        & ODC$_{TrackInit}$ & \textbf{24.86} & 38.76 & 40.00 & 37.88 & 31.32\\
        & ODCT & 24.58 & \textbf{39.74} & \textbf{42.80} & \textbf{43.83} & \textbf{38.28}\\
        \hline
    \multirow{3}{*}{\textit{K = 300}}
        & ODC & 24.01 & 38.67 & 39.57 & 37.55 & 31.62\\
        & ODC$_{TrackInit}$ & \textbf{24.87} & 38.85 & 41.47 & 38.70 & 33.10\\
        & ODCT & 24.77 & \textbf{41.00} & \textbf{44.56} & \textbf{45.55} & \textbf{41.38}\\
        \hline
    \multirow{3}{*}{\textit{K = 1000}}
        & ODC & 24.22 & 38.36 & 39.56 & 37.57 & 31.58\\
        & ODC$_{TrackInit}$ & 24.74 & 39.63 & 40.50 & 38.77 & 33.37\\
        & ODCT & \textbf{25.08} & \textbf{41.28} & \textbf{44.57} & \textbf{46.23} & \textbf{41.74} \\
        
    \hline
    \end{tabular}
    }
    \end{minipage} 
\end{table*}

\subsection{Preliminary evaluation}
\label{preliminary}
In a preliminary set of experiments, we analyze the models' behavior during training. 

\subsubsection{Cluster reassignments}
To assess the models' stability, we keep track of the change ratio, i.e. the fraction of samples whose labels change at each iteration. Intuitively, fewer label switches indicate better stability.
In Fig. \ref{fig:changeratio} we illustrate the change ratio during training for ODC, ODC$_{TrackInit}$ and ODCT, trained from scratch, with $K=300$ clusters. 
Models with $K=30, 1000$ presented similar trends.
Initially, almost 100\% of samples undergo a label switch. The ratio decreases gradually and converges to $\sim 0.1$, thus indicating stability for 9 samples out of 10.
Interestingly, ODC$_{TrackInit}$ and ODC have almost identical curves, while ODCT initially exhibits a higher change ratio but is able to eventually converge to a more stable solution ($\sim 0.04$). 


\subsubsection{Cluster entropy}
\label{sec:entropy}
To assess how well the models assigns samples from the same track to the same cluster, we rely on an entropy based evaluation.
Given a cluster assignment, for each track $T$ we can determine the sample distribution over the $K$ clusters, counting the occurrences of track samples per each cluster, $X_T=\{x_1, x_2, ..., x_K\}$.
We then compute the Shannon entropy $H$, for $X_T$, as 
    $H(X_T)=-\sum^K_{i=1}{x_ilog(x_i)}$.
This measures how noisy the distribution of track samples is over clusters. Ideally, samples of a same track should belong to the same cluster, yielding entropy $H(X_T)=0$.

In Fig. \ref{fig:meantrackentropy} we report the mean track entropy computed for all the ImageNet VID track proposals during training.
At the first epoch, we have $H(X_T)=0$ for ODC$_{TrackInit}$ and ODCT since all the samples in a track are associated to the same cluster and same pseudo-label.
ODC$_{TrackInit}$ mostly follows the same trend as ODC, meaning that initialization alone does not suffice to obtain an effective clustering during training. Nonetheless, it still obtains a slightly lower entropy value than vanilla ODC, indicating that a good initialization indeed has a beneficial effect.
ODCT, on the other hand, due to the constrained Samples Memory update, exhibits a completely different trend, managing to effectively lower the mean track entropy epoch by epoch.

\subsection{Transfer Learning Experiments}
\label{transfer}
As in \cite{zhang2017splitbrain}, we extract image features from our ResNet-50 pretrained on Imagenet VID, truncating the model after the last layer of every residual stage in ResNet-50. 
In the following we discuss the downstream tasks we adopted to evaluate our pre-trained model and the quality of such features.

			
			

\subsubsection{ImageNet Linear Classification}
\label{sec:expimagenet}

As proposed by Zhang \textit{et al.} in \cite{zhang2016colorful}, 
we test the task generalization of the representation by freezing the backbone's weights and training linear classifiers on top of each layer to perform 1000-way ImageNet classification.
We report top-1 center-crop accuracy on the validation split of ImageNet. Results are shown in Tab. \ref{tab:imagenetpascal} (left).
We compare our model against ResNet-50 trained directly on ImageNet in a fully supervised setting and againts a version with random weights, as proposed in \cite{articlek}. Such methods act as an upper bound and lower bound for the task, respectively.
We also report results for ODC pretrained on ImageNet, as reference.
We compare our model to ODC pretrained on ImageNet VID for different number of clusters values.
All models are trained for 100 epochs, using SGD with momentum of 0.9 and batch size of 64. The learning rate is initialized as 0.01 and decayed by a factor of 10 after every 30 epochs.

We can observe that ODCT Stage1 results in slightly higher accuracy than ODC and ODC$_{TrackInit}$, with the best improvement using $K=300$.
However, this small performance gap is immediately widened at Stage2, and our ODCT network achieves much higher results compared to standard ODC and ODC$_{TrackInit}$ pre-trained on ImageNet VID. As we use features extracted from deeper layers of the network, we can observe a consistent increase up to Stage4 for all methods. In particular, ODTC exhibits and improvement across all layers, with the best gain ($\sim$33\%-40\%) obtained with Stage4 and Stage5 and $K=300, 1000$.
These results indicate that exploiting video track information encourages representations that linearly separate semantic classes in the trained data distribution.

However, there is no significant improvement using $K=1000$ compared to $K=300$. This suggests that, after a certain value of $K$, the accuracy saturates.
Interestingly, we can see that also for ODC$_{TrackInit}$ there is a considerable performance gain compared to vanilla ODC, proving that a good initialization can have a significant impact on the whole training process.

\begin{figure*}
	\centering
	\includegraphics[width=.32\textwidth]{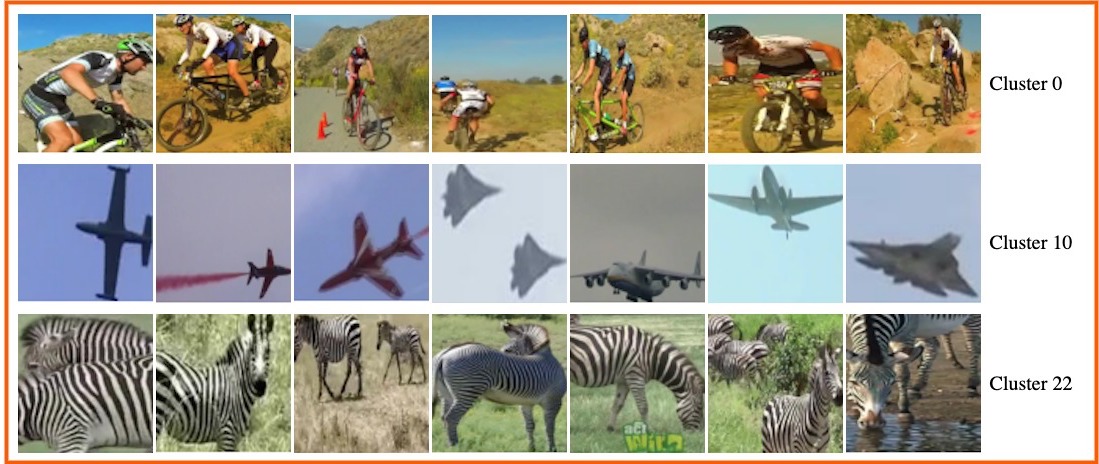}
	\includegraphics[width=.32\textwidth]{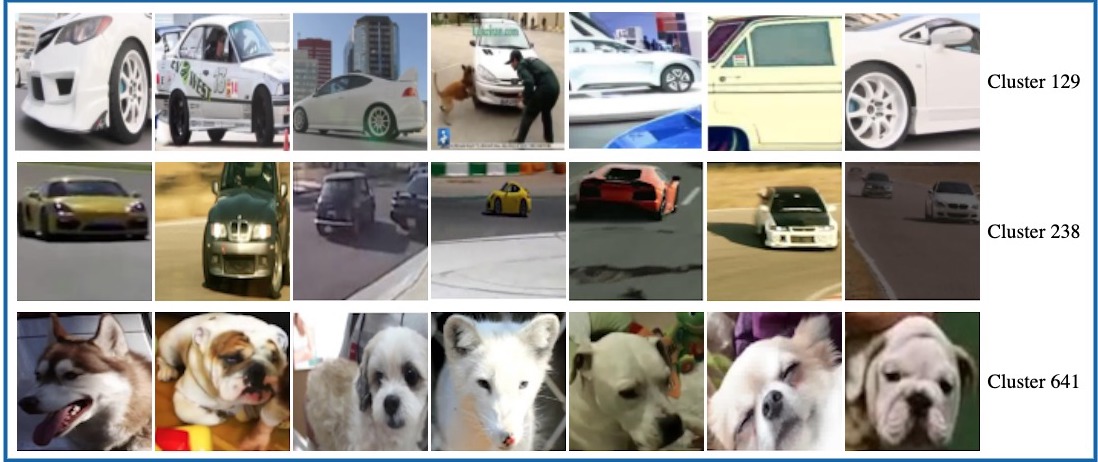}
	\includegraphics[width=.32\textwidth]{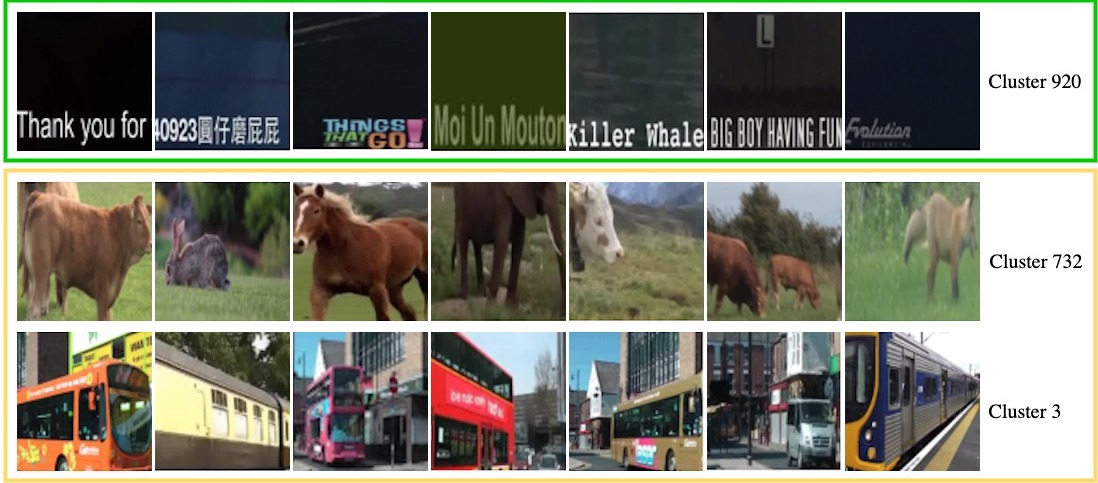}
	\caption{Samples assigned to different clusters. Orange: existing classes. Blue: subclasses. Green: new classes. Yellow: object relations.}
	\label{fig:cluster}
\end{figure*}
	
\subsubsection{VOC2007 SVM Classification}
\label{svmtask}

We test our model on the classification benchmark PASCAL VOC2007\cite{pascal-voc}.
Following the setup of Goyal \textit{et al.} \cite{goyal2019scaling}, we train linear SVMs on features extracted from the ResNet-50 backbone on the \texttt{trainval} split of VOC07. 
We follow the same configuration of \cite{goyal2019scaling} and report mean Average Precision (mAP) on the test set.
The results in Tab. \ref{tab:imagenetpascal} (right), show that ODCT surpasses ODC, when both are pre-trained on ImageNet VID, by a significant margin on the VOC2007 SVM classification task.

We can observe that, ODCT yields comparable results to ODC$_{TrackInit}$ for $K=30, 300$ for Stage1. However, starting from Stage2, ODCT achieves higher results throughout the remainder of the network compared to other unsupervised methods pretrained on ImageNet VID.
Significantly, with $K=1000$, ODCT achieves 46.23\% mAP with Stage4, which is 6,67\% higher than the best performing layer of ODC and 5,73\% higher than the best performing layer of ODC$_{TrackInit}$.
Also, our method gains over 10\% mAP points with $K=300, 1000$ compared to ODC on the last layer.
	
\subsection{Qualitative Analysis}
In Fig. \ref{fig:cluster} we show portions of the clusters obtained training ODCT using $K=1000$.
The number of clusters is much larger than that of the annotated classes and interestingly there are some clusters that represent new semantic categories beyond the annotated ones.
In addition to the clusters that represent existing classes in ImageNet VID, shown in the orange box, we find additional subclasses (blue box), e.g. "white car", "car on racing track" or "dog with white snoug".

ODCT also groups images with similar relations between objects. As shown in the yellow box, the method discovers clusters representing “animal on a lawn” and "busses and trains", which have a similar visual appearance.
Moreover, due to the unsupervised pre-training on videos, ODCT is capable to detect new classes such as "text on a dark background" (green box in Fig. \ref{fig:cluster}).

	\begin{table}[t]
	\caption{Intra-Track and Intra-Class Entropy computed with GT tracks from ImageNet VID. The lower the better.}
	\label{tab:entropy}
			\begin{tabular}{llcc}
				\hline
				& \textbf{Method} & \textbf{Intra-Track H} & \textbf{Intra-Class H} \\ 
				\hline
				
				\multirow{3}{*}{\textit{K = 30}} & ODC$_{GT}$ & 0.868 & 2.986\\
				
				& ODC$_{GT-TrackInit}$ & 0.873 & 3.013\\
				
				& ODCT$_{GT}$ & \textbf{0.061} & \textbf{2.813}\\
				
				\hline
				\multirow{3}{*}{\textit{K = 300}} & ODC$_{GT}$ & 1.368 & 4.963\\
				& ODC$_{GT-TrackInit}$ & 1.360 & 4.969\\
				& ODCT$_{GT}$ & \textbf{0.064} & \textbf{4.099}\\
				
			\end{tabular}
	\end{table}

\subsection{Training with Ground Truth Tracks}
\label{sec:gt}
Previous experiments demonstrated that ODCT is effective and achieves a significant improvement in transfer learning tasks when compared to vanilla ODC. Such results are obtained relying on a set of tracks generated without supervision, as outlined in Sec. \ref{sec:track}.
Since ImageNet VID also contains manually annotated tracks, we retrain our model using GT tracks (discarding class information) to highlight the difference between the two approaches.
Our unsupervised track generation method yields a total of 57,879 tracks compared to the 7,857 ground truth tracks annotated in the dataset. Our tracks however, despite being an order or magnitude more, are likely to contain noise, be fragmented or focus on object parts or groups of objects. On the other hand, GT tracks are precise, clean and represent single objects in their entirety.

Following the same setting described in section \ref{sec:trainingdetails} we train from scratch the three models on the GT ImageNet VID tracks\footnote{We use 10\% of equally spaced samples per track, a total of 77,714 samples.}, with $K = 30, 300$, discarding class labels. We refer to these models as ODC$_{GT}$, ODC$_{GT-TrackInit}$ and ODCT$_{GT}$.

We evaluate the models using two entropy measures:

(i) \textit{Intra-Track Entropy} is used to evaluate the distribution of tracks over clusters. For each track $T$, we measure how a track is distributed over $K$ clusters by computing the Shannon entropy as in Sec. \ref{sec:entropy}, counting the occurrences of track samples per each cluster.

(ii) \textit{Intra-Class Entropy} is a measure of how well a model assigns samples from the same class to the same cluster. For each class, we compute the class distribution over $K$ clusters, counting the occurrences of class samples per each cluster, and then compute the Shannon entropy.

In Tab. \ref{tab:entropy} we report the Intra-Track and Intra-Class entropies computed on all tracks.
In both the experiments, the entropy for ODCT is lower, thus indicating that our approach assigns samples from the same tracks and classes to the same clusters better than the others. In particular, it is interesting to notice the significant drop in Intra-Track entropy compared to the other methods. This underlines the effectiveness of our clustering strategy, which manages to keep together samples belonging to the same track.
	
\begin{table}[t]
\caption{ImageNet top-1 accuracy and Pascal VOC07 mAP for ODCT trained on unsupervised tracks and GT tracks.}
\label{tab:gt}
\centering
	\scriptsize{
		\begin{tabular}{llcc}
			\hline
			& \textbf{Method} & \textbf{ImageNet top-1 acc} & \textbf{VOC07 mAP} \\ 
			\hline
			
			\multirow{3}{1cm}{\textit{K = 30}}
			& ODC & 22.86 & 36.72 \\
			& ODC$_{TrackInit}$ & 24.07 & 37.88 \\
			& ODCT & \textbf{30.26} & \textbf{43.83} \\
			\hdashline
			\multirow{3}{*}{\textit{K = 30}}
			& ODC$_{GT}$ & 19.75 & 33.77 \\
			& ODC$_{GT-TrackInit}$ & 20.01 & 33.36 \\
			& ODCT$_{GT}$ & \textbf{22.49} & \textbf{36.14}\\
			\hline
			\multirow{3}{1cm}{\textit{K = 300}}
			& ODC & 23.35 & 37.55 \\
			& ODC$_{TrackInit}$ & 25.86 & 38.70 \\
			& ODCT & \textbf{32.23} & \textbf{45.55} \\
			\hdashline
			\multirow{3}{*}{\textit{K = 300}}
			& ODC$_{GT}$ & 21.33 & 35.25 \\
			& ODC$_{GT-TrackInit}$ & 21.42 & 35.59 \\
			& ODCT$_{GT}$ & \textbf{24.18} & \textbf{37.29} \\
		\end{tabular}
	}
\end{table}

We additionally evaluate the effectiveness of the model trained on the GT tracks using the ImageNet and VOC07 downstream classification tasks. We use the same settings indicated in Sec. \ref{sec:expimagenet} and Sec. \ref{svmtask}, considering only Stage4. Results are shown in Tab. \ref{tab:gt}.
Interestingly, there is not much difference among all methods using the ground truth annotations, with a gap of at most 2-3 points in either accuracy and mAP.
At the same time, it is surprising to notice that unsupervised tracks yield much higher results when comparing models to their counterparts trained with GT tracks.
This hints to the fact that, even if manually annotated tracks are clean and precise, it is better to train with more, possibly noisy, data. In addition, our unsupervised track generation can provide tracks from unseen classes that can help to perform a better pre-training.

\section{Conclusions}

In this paper we have introduced an unsupervised clustering-based approach that exploits temporal consistency of video tracks to model intra-class variation. It emerges that simple track based initialization has beneficial effects and overall exploiting temporal information leads to important gains in accuracy compared to prior work based on still images.






\bibliographystyle{IEEEtran}
\bibliography{ref.bib}
%
%
%

\end{document}